\pdfoutput=1

\documentclass[11pt]{article}

\usepackage[final]{acl}

\usepackage{times}
\usepackage{latexsym}
\usepackage{graphicx}
\usepackage{booktabs}
\usepackage{amsmath}
\usepackage[T1]{fontenc}
\newcommand{\tabitem}{~~\llap{\textbullet}~~}
\usepackage[utf8]{inputenc}
\usepackage{enumitem}
\usepackage{multirow}
\usepackage{bbding}
\usepackage{microtype}
\usepackage{tcolorbox}
\usepackage{graphicx}
\usepackage{tikz}
\usepackage{forest}
\def\halfcheckmark{\tikz\draw[scale=0.4,fill=black](0,.35) -- (.25,0) -- (1,.7) -- (.25,.15) -- cycle (0.75,0.2) -- (0.77,0.2)  -- (0.6,0.7) -- cycle;}
\usetikzlibrary{trees,positioning,shapes,shadows,arrows.meta}
\title{How to Enable Effective Cooperation Between Humans and NLP Models: \\A Survey of Principles, Formalizations, and Beyond}

\author{Chen Huang$^{\spadesuit\diamondsuit}$, Yang Deng$^{\flat}$, Wenqiang Lei$^{\spadesuit\diamondsuit}$\thanks{Correspondence to Wenqiang Lei.}, \\\textbf{Jiancheng Lv}$^{\spadesuit\diamondsuit}$, \textbf{Tat-Seng Chua}$^{\heartsuit}$, \textbf{Jimmy Xiangji Huang}$^{\clubsuit}$ \\
 ${\spadesuit}$ Sichuan University \quad $\flat$ Singapore Management University \quad ${\clubsuit}$York University \\ 
 ${\diamondsuit}$ Engineering Research Center of Machine Learning and Industry Intelligence, \\Ministry of Education, China \quad ${\heartsuit}$ National University of Singapore  \\
  \texttt{\{huangc.scu, dengyang17dydy\}@gmail.com, wenqianglei@scu.edu.cn}\\ \texttt{lvjiancheng@scu.edu.cn, chuats@comp.nus.edu.sg, jhuang@yorku.ca}
  }

\begin{document}
\maketitle

\begin{abstract}
With the advancement of large language models (LLMs), intelligent models have evolved from mere tools to autonomous agents with their own goals and strategies for cooperating with humans. This evolution has birthed a novel paradigm in NLP, i.e., human-model cooperation, that has yielded remarkable progress in numerous NLP tasks in recent years. In this paper, we take the first step to present a thorough review of human-model cooperation, exploring its principles, formalizations, and open challenges. In particular, we introduce a new taxonomy that provides a unified perspective to summarize existing approaches. Also, we discuss potential frontier areas and their corresponding challenges. We regard our work as an entry point, paving the way for more breakthrough research in this regard.

\end{abstract}

\section{Introduction}

Advancements in NLP research have been greatly propelled by large language models (LLMs), which have showcased exceptional abilities \cite{zhao2023survey, laskar2024systematic}. These advancements are paving the way for the development of AI models that can behave as autonomous agents, working alongside humans to tackle intricate tasks. 
These models, for example, can cooperate with humans on data annotation \cite{klie-etal-2020-zero, li2023coannotating, huang2024araida}, information seeking \cite{deng2023survey, wang2023rethinking, zhang2024clamber}, creative writing \cite{padmakumar2021machine, akoury2020storium} and real-world problem solving \cite{mehta2023improving, feng2024large, qian2024tell}.
This growing synergy between humans and models has fueled a surge of research into a new paradigm: \textbf{Human-Model Cooperation}. This paradigm, facilitated by diverse user interfaces ranging from natural language conversations \cite{ni2023recent} to action sequences like clicking buttons \cite{chen-etal-2023-travel, rosset2020leading}, holds promise for unlocking unprecedented levels of efficiency across various domains.


In fact, establishing intelligent models that can interact with humans has always been a longstanding research \cite{press1971toward, wallenius1975comparative, milewski1997delegating, dzindolet2003role, bahner2008misuse, chien2018attention, touvron2023llama, achiam2023gpt}. The realm of NLP tasks has seen a surge in methods for human-model cooperation \cite{wang2023interactive}, particularly in the era of LLMs and agents \cite{xi2023rise}. Recently, emergent surveys also have made commendable strides \cite{wang2021putting, wang2023interactive, wu2023designing, yang2024human, gao2024taxonomy}. However, they primarily focus on introducing key elements of human-model cooperation, such as user interfaces (e.g., dialogues), message understanding and fusion, cooperation system evaluation, and applications to NLP tasks (see Table \ref{tab:related} for details). Given all these elements, \textbf{the information on particular details about how to formalize an effective human-model cooperation to achieve collective outputs is rather under-specified and scattered}. Therefore, a comprehensive and systematic analysis of the underlying principles and formalizations of human-model cooperation is still absent. This gap in understanding presents a significant opportunity for advancement, enabling us to develop a deeper understanding of the fundamental basics that govern the effective cooperation between humans and intelligent models.

\begin{table*}[]
\centering
\setlength{\abovecaptionskip}{5pt}   
\setlength{\belowcaptionskip}{0pt}
\resizebox{0.99\textwidth}{!}{%
\begin{tabular}{l|l|c|c|c|p{8cm}}
\toprule
\multicolumn{2}{l|}{\textbf{Cooperation Formalization}} & \multicolumn{1}{c|}{\textbf{\begin{tabular}[c]{@{}c@{}}Who Make\\ Final Decision\end{tabular}}} & \multicolumn{1}{c|}{\textbf{\begin{tabular}[c]{@{}c@{}}Role\\ Framework\end{tabular}}} & \multicolumn{1}{c|}{\textbf{\begin{tabular}[c]{@{}c@{}}Decision-making\\ Independently\end{tabular}}} & \multicolumn{1}{c}{\textbf{\begin{tabular}[c]{@{}c@{}}Representative Methods for Different Categories \\
(Details are presented in Appendix \ref{app:overview})
\end{tabular}}} \\ \midrule
\multicolumn{2}{l|}{\begin{tabular}[c]{@{}l@{}}
\textbf{Sequential Cooperation}, \\
where two parties work together \\
in a step-by-step manner, \\
with each step building \\
upon the previous one  (Sec. \ref{seq}) \end{tabular}} & Human or Model & Assistor-Executor & No & \begin{tabular}[c]{@{}p{8cm}@{}} \tabitem Human-assisted method \cite{liu2018dialogue, santurkar2021editing, touvron2023llama, mehta2023improving, wang2023mint} \\ \tabitem Model-assisted method \cite{lai2019human, alslaity2019towards, li2021fitannotator, donahue2022pick, huang2024araida} \end{tabular} \\ \midrule
\multicolumn{2}{l|}{\begin{tabular}[c]{@{}l@{}}
\textbf{Triage-based Cooperation}, \\
where tasks/data are strategically \\
distributed between two parties \\ (Sec. \ref{tri}) \end{tabular}} & Human or Model & Equal-Partnership & Yes & \begin{tabular}[c]{@{}p{8cm}@{}} \tabitem Model-based allocator \cite{thulasidasan2019combating, mozannar2020consistent, deng-etal-2022-pacific} \\ \tabitem Extra allocator \cite{wang2021classification, huang-etal-2023-reduce, li2023coannotating, huang2024selective} \end{tabular} \\ \midrule
\multicolumn{2}{l|}{\begin{tabular}[c]{@{}l@{}}
\textbf{Joint Cooperation}, \\
where the final outcome \\
resulting from the collective \\
decisions of two parties (Sec. \ref{joi})\end{tabular}} & Human and Model & Equal-Partnership & No & \begin{tabular}[c]{@{}p{8cm}@{}}\tabitem Probabilistic approach \cite{kerrigan2021combining, huang2024comatching} \end{tabular} \\ \bottomrule
\end{tabular}%
}
\caption{Overview of Human-Model Cooperation. Our taxonomy is based on how cooperation takes place and who ultimately takes responsibility for the final decision, identifying three main types of cooperation. For better understanding, we showcase typical applications of human-model cooperation in Appendix \ref{applica}.}
\label{tab:myfea}
\vspace{-3mm}
\end{table*}

To fill this gap, in this survey, we take the first step to summarize the principles, formalizations, and open challenges of human-model cooperation\footnote{Refer to Appendix \ref{liter} for our literature review process}. 
We begin by introducing the definition and principles of this rapidly evolving field, providing a common ground for understanding. Next, we propose a new and systematic taxonomy for cooperation formalizations, offering a unified perspective to summarize existing approaches. This taxonomy, based on how cooperation takes place and who ultimately bears responsibility for decision-making, identifies three distinct types of cooperation, each with its own unique role framework defining the contributions of both cooperators to the overall task.
Finally, we delve into potential research frontiers, highlighting both technical considerations and social impact. These frontiers identify opportunities and challenges for future investigation, paving the way for more advancements. 
As such, this survey seeks to stimulate further research and advance our understanding of human-model cooperation.
This understanding is vital for harnessing the full potential of the cooperators and shaping a future where humans and intelligent models work together seamlessly. Our major contributions are as follows:
\begin{itemize}[leftmargin=*, itemsep=-4pt]
    \item For the first time, we provide a comprehensive survey of the principles and formalizations of human-model cooperation.
    \item We introduce a novel and systematic taxonomy that offers a unified perspective on existing approaches to formalizing human-model cooperation, as illustrated in Table \ref{tab:myfea}.
    \item We identify key research frontiers and their associated challenges, paving the way for groundbreaking research that will advance the field of human-model cooperation.
\end{itemize}

\section{Definitions \& Principles of Human-Model Cooperation}
\textbf{Definitions}. Over the past few decades, various terms have been used to depict the concept of human-model cooperation. These terms often carry comparable meanings and are occasionally interchangeable. To address this issue, we establish clear definitions for human-model cooperation, carefully differentiating it from other terms. This provides a starting point for exploring this field.

\begin{center}
\begin{tcolorbox}[colback=gray!10,
                  colframe=black,
                  width=7.7cm,
                  arc=1mm, auto outer arc,
                  boxrule=0.5pt,
                  left=2pt, right=2pt, top=2pt, bottom=2pt
                 ]
\textbf{Human-Model Cooperation} involves the human and the model working together as a unified team, engaging in the decision-making process of shared tasks to achieve a \textit{shared goal}.
\end{tcolorbox}
\end{center}

Unlike non-cooperation \cite{deng2023survey}, shared or aligned goals form the foundation for effective human-model cooperation \cite{jiang2022computational, wang2021putting}. However, cooperation doesn't always mean sharing resources or information. Two parties can work independently to achieve a shared goal without mutual communication. However, \textit{human-model collaboration} can go beyond cooperation \cite{hord1981working}. It involves an equal partnership, where both parties work together through bidirectional communication, shared decision-making, and interdependence. This often results in a more coordinated and efficient approach to achieving a shared goal. We focus on the cooperation, leaving discussions on collaboration for future work (cf. Section \ref{build}).

\begin{figure*}[!htp]
    \centering
    \setlength{\abovecaptionskip}{1pt}   
    \setlength{\belowcaptionskip}{1pt}
    \includegraphics[width=0.85\textwidth]{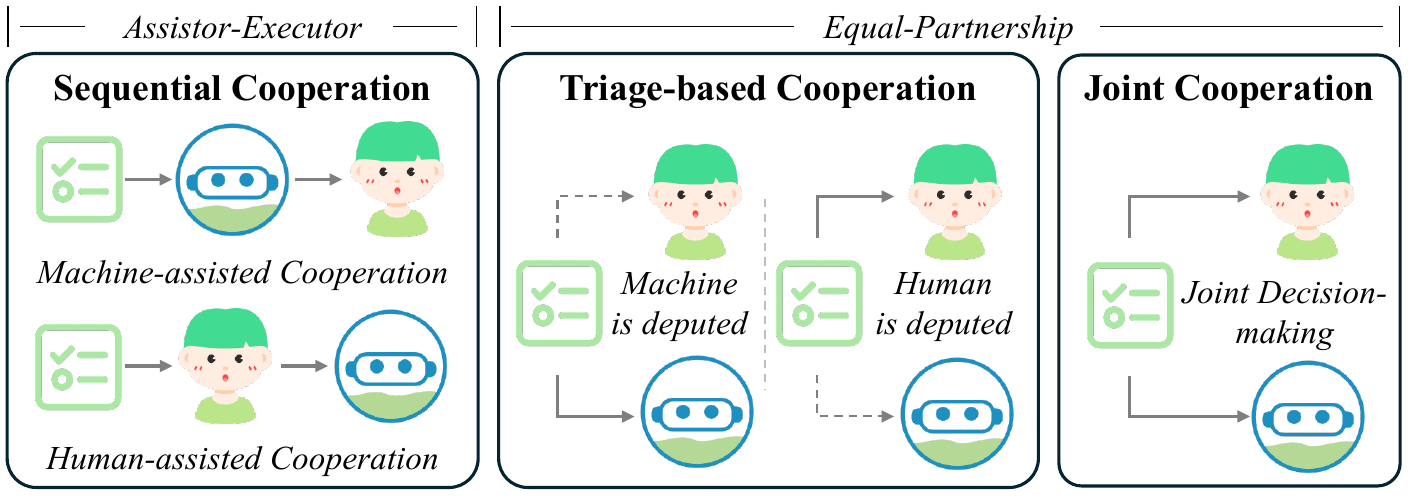}
    \caption{Unified taxonomy for categorizing the formalization of human-model cooperation. We also introduce two role frameworks, defining how the cooperators contribute to the overall task.}
    \label{fig:enterddd-label}
    \vspace{-3mm}
\end{figure*}

\begin{table}[t]
\centering
\setlength{\abovecaptionskip}{5pt}   
\setlength{\belowcaptionskip}{0pt}
\resizebox{0.45\textwidth}{!}{%
\begin{tabular}{l|ccc|l}
\toprule
\textbf{Surveys} & \textbf{EC} & \textbf{FC} & \textbf{PC} & \begin{tabular}[c]{@{}l@{}}\textbf{Remarks}\end{tabular} \\ \midrule
\citet{wang2021putting} & \Checkmark & \textbackslash{} & \textbackslash{} & \begin{tabular}[c]{@{}l@{}}Human-model NLP tasks, \\ Interation objectives, \\ Types of human feedback, \\ User interfaces\end{tabular} \\ \hline
\citet{gao2024taxonomy} & \Checkmark & \textbackslash{} & \textbackslash{} & \begin{tabular}[c]{@{}l@{}}Interaction modes \\(Prompting, User Interface, \\Context, and Agent Facilitator)\end{tabular}\\ \hline
\citet{wang2023interactive} & \Checkmark & \textbackslash{} & \textbackslash{} & \begin{tabular}[c]{@{}l@{}}Interactive objects, \\ User interfaces, \\ Message fusion strategies, \\ Cooperation system evaluation\end{tabular} \\  \hline
\begin{tabular}[c]{@{}l@{}}\citet{wu2023designing}\\ \citet{yang2024human}\end{tabular} & \Checkmark & \textbackslash{} & \textbackslash{} & \begin{tabular}[c]{@{}l@{}}Cooperation system evaluation,\\ User interfaces, \\ Learn from human feedback\end{tabular} \\ \hline
\citet{xi2023rise} & \Checkmark & \textbackslash{} & \textbackslash{} & \begin{tabular}[c]{@{}l@{}}Cooperator role\\ of human-agent cooperation\end{tabular} \\ \hline
\citet{wan2022user} & \textbackslash{} & \halfcheckmark & \textbackslash{} & \begin{tabular}[c]{@{}l@{}}
Detailed categories for \\
one specific type \\
of cooperation form\end{tabular} \\ \midrule
\textbf{Ours} & \textbackslash{} & \Checkmark & \Checkmark & \begin{tabular}[c]{@{}l@{}}Cooperation principles,\\ Taxonomy of cooperation\\ forms (three types)\end{tabular} \\ \bottomrule
\end{tabular}
}
\caption{Our differences to related surveys. `\textbf{EC}', `\textbf{FC}', and `\textbf{PC}' refers to elements, formalizations, and principles of cooperation, respectively. A systematic analysis of the underlying principles and formalizations of human-model cooperation is still absent. Appendix \ref{related} details our distinctions from related surveys.}
\label{tab:related}
\vspace{-3mm}
\end{table}


\textbf{Principles}. While many factors can affect how both parties behave, rational individuals usually aim to align their actions with principles to ensure effective and meaningful cooperation.
Drawing inspiration from foundational work in conversational theory, our survey reinterprets the cooperative principles outlined by \citet{grice1975logic, grice1989studies} to broaden their applicability beyond cooperative conversation and extend them to general cooperative applications. As evidenced by early studies \cite{grice1975logic, grice1989studies, sari2020flouting}, the cooperative principle can be split into the following four maxims\footnote{Maxim by definition is a phrase or saying that includes a rule or moral principle about how one should behave.}:
\begin{itemize}[leftmargin=*, itemsep=-4pt]
    \item \textbf{Sincerity}—Do what one believes to be true. Each participant should act sincerely without deception and ensure that their responses are backed by sufficient evidence. Taking the conversational recommendation systems (CRS) as an example, it should provide credible recommendation explanations \cite{qin-etal-2024-beyond}.
    \item \textbf{Relation}—Make one's contribution contribute to solving the goals of the task. The actions of the participant need to be task-relevant or cooperation-relevant. For instance, the responses from CRS should contribute to identifying user preferences and making recommendations.
    \item \textbf{Manner}—Make one's contribution appropriate in complexity to the requirements of the task goals. Each participant's responses should be easily understood and clearly expressed. In this case, CRS's responses should be lucid.
    \item \textbf{Quantity}—Make one's contribution as informative as required for the task goals. Each participant should provide the necessary level of information without overwhelming the other with unnecessary details. In this case, recommendation explanations from CRS should avoid being lengthy and repetitive.
\end{itemize}



\section{Formalization Taxonomy}
\textbf{Overview}. We provide a fine-grained and unified taxonomy for categorizing the formalization of human-model cooperation methods, as illustrated in Figure \ref{fig:enterddd-label} and Table \ref{tab:myfea}. Our taxonomy is based on who ultimately takes responsibility for the final decision, identifying three main types of cooperation, including 1) \textit{sequential cooperation}, 2) \textit{triage-based cooperation}, and 3) \textit{joint cooperation}. Each form of cooperation adheres to a specific role framework, defining how the cooperators contribute to the overall task. We also showcase typical applications of human-model cooperation in Appendix \ref{applica} for better understanding.

\textbf{Adherence to Principles}. Guided by the cooperative principle, existing methods for human-model cooperation fundamentally rely on the assumption that both parties act rationally. This implies that they try to take the best action toward achieving their goals instead of making decisions randomly and maliciously. For example, existing methods often rely on fully cooperative user simulators when conducting experiments on human-machine cooperation. These simulators typically define the specifics of cooperation during interaction with the model through rules \cite{zhang2020evaluating, lei2020estimation} or prompts \cite{sekulic2022evaluating, wang2023rethinking}, such as instructing the simulator to provide truthful answers to model's clarifying questions \cite{zhang2024clamber}. Further details on the specific type of cooperation will be provided in the corresponding subsection.

\textbf{Role Frameworks}. Human-model cooperation encompasses a diverse range of roles, leading to two role frameworks \cite{zhang2021ideal, xi2023rise}: the assistor-executor framework and the equal-partnership framework. These frameworks are differentiated by the degree of shared responsibility for the final decision.
\begin{itemize}[leftmargin=*, itemsep=-4pt]
    \item \textbf{Assistor-executor framework}. This framework, the most prevalent one, reflects a hierarchical approach where one party holds primary decision-making power. In particular, one party acts as the assistant, providing guidance and information, while the other, the executor, retains the authority to make the final decision. 
    \item \textbf{Equal-partnership framework}. Both parties contribute equally to decision-making in the equal-partnership framework, participating on an equal footing with two parties in cooperation. This framework may promote a more democratic approach to human-model interaction. However, regulation and accountability towards the human-model cooperation system become more complex in this framework as both parties share decision-making power (cf. Section \ref{use} for discussion). 
\end{itemize}

\subsection{Sequential Cooperation}
\label{seq}
\begin{center}
\begin{tcolorbox}[colback=gray!10,
                  colframe=black,
                  width=7.7cm,
                  arc=1mm, auto outer arc,
                  boxrule=0.5pt,
                  left=2pt, right=2pt, top=2pt, bottom=2pt
                 ]
\textbf{Sequential Cooperation} refers to a cooperative process where the human and the model work together in a step-by-step manner, with each step building upon the previous one.
\end{tcolorbox}
\end{center}

\textbf{Overview}. Sequential cooperation is the most prevalent form in NLP \cite{wang2021putting, wan2022user}. In this context, the actions or decisions of one party are influenced by the actions or decisions of the other participant in a sequential fashion \cite{brocas2018path}, primarily concerning increased efficiency and agency \cite{sperrle2021survey}. Therefore, this type of cooperation often involves a series of interdependent tasks or actions that are carried out in a specific order to achieve a common goal or outcome. Notably, sequential cooperation mirrors the hierarchical structure of the assistor-executor framework. The first party in this sequence typically assumes the role of the assistant, providing suggestions or feedback that adhere to cooperative principles, while the second party, the executor, ultimately takes responsibility for the final decision. Due to the straightforward nature of this cooperation form, the NLP community has enthusiastically adopted this approach, achieving significant successes in various applications, including data collection and annotation \cite{fanton-etal-2021-human, klie-etal-2020-zero, casanova2020reinforced, li2021fitannotator}, technology-assisted manual document review workflows \cite{10.1145/3459637.3482415,10.1145/2600428.2609601, 10.1145/2911451.2911510}, conversational information retrieval \cite{zamani2022conversational}, schema induction \cite{zhang-etal-2023-human}, fact checking \cite{mendes-etal-2023-human}, creative writing and summarization \cite{chen-etal-2023-human, padmakumar2021machine, akoury2020storium}, drug editing \cite{liu2023chatgptpowered}, cooperative agent \cite{zhang2024proagent}, and many others \cite{gerlach2021multilingual, sharma2023human}.

\begin{figure}
    \centering
    \setlength{\abovecaptionskip}{1pt}   
    \setlength{\belowcaptionskip}{1pt}
    \includegraphics[width=0.34\textwidth]{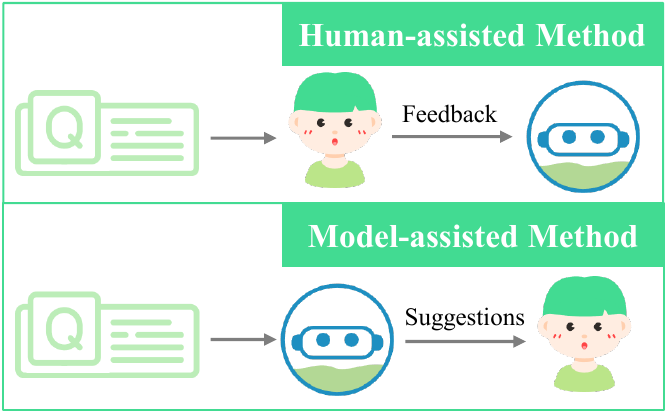}
    \caption{Two types of sequential cooperation based on whether the model or the human, respectively, takes on the role of the assistor.}
    \label{fig:mac}
    \vspace{-3mm}
\end{figure}

\textbf{Methods}. Given the widespread use of sequential cooperation, we delve deeper into its two specific types: \textit{model-assisted} and \textit{human-assisted method}, as illustrated in Figure \ref{fig:mac}. These distinctions are based on whether the model or the human, respectively, takes on the role of the assistor.
\begin{itemize}[leftmargin=*, itemsep=-4pt]
    \item \textbf{Human-assisted Method}\footnote{Also called \textit{human-in-the-loop}.} involves the model assuming the ultimate decision-making authority while leveraging human expertise to enhance its capabilities by providing feedback\footnote{Current methods operate under the assumption that human feedback is always accurate and should be accepted.} on the predicted or intermediate model results. This human feedback could extend throughout the model's lifecycle – \textit{from data processing and model selection to optimization, alignment, and evaluation} – with human assistance/feedback continuously shaping the model's development \cite{wang2021putting, touvron2023llama}. As a result, human-assisted method can lead to more personalized models \cite{bae2020interactive, liu2020learning} and mitigate potential biases and errors inherent in automated processes \cite{mosqueira2023human, bai2022training, fails2003interactive, zhang2025enhancingcodellmtraining}. To achieve this, existing research can be categorized into two groups based on how to learn from human feedback: 1) \textit{Training-based approach} translates human feedback into supervisions, which are used to train the task-specific model in either an offline \cite{qian2024tell, touvron2023llama} or online fashion \cite{liu2018dialogue, kumar2019didn}; 2) \textit{Training-free approach} resort to the in-context learning \cite{huang2022inner, wang2023mint, cai2023human, wu2022ai}, model editing \cite{santurkar2021editing, huang2024dreditor, cheng2024editing}, and rule learning \cite{yang2019study} as more efficient alternatives to learn from human feedback. 
    \item \textbf{Model-assisted Method}\footnote{Also called \textit{algorithm-in-the-loop} or \textit{machine-in-the-loop}.} enhances human decision-making by leveraging model assistance \cite{green2020algorithm, lai2021towards, punzitowards}. In this scenario, the two parties work autonomously, with the model performing specific tasks to assist the human's duty and provide candidate solutions. The human, in turn, monitors the model's work and ultimately makes the final decision based on the solutions. In practice, the model either proposes a single solution \cite{li2021fitannotator, sharma2023human}, which the human can then accept or reject\footnote{If rejected, the human can modify the solution}, or presents top-$k$ solutions to to narrow down the possibilities, which the human can then either refine these solutions into a new one \cite{memmert2023towards} or directly select the final solution from the presented list \cite{donahue2022pick, straitouri2023improving, kilgarriff2008gdex}. As a result, compared to humans completing tasks from scratch, this form of cooperation is believed to significantly reduce the human workload. However, its effectiveness hings on both the accuracy and trustworthiness of model-generated solutions, encouraging human acceptance and ultimately minimizing workload while maximizing efficiency. To achieve this, there are two primary ways: 1) \textit{Suggesting accurate solutions}. Researchers have employed various techniques, apart from using task-specific LMs/LLMs, including incorporating active learning strategies \cite{li2021fitannotator, klie-etal-2018-inception}, analogical reasoning \cite{huang2024araida}, and conformal prediction methods \cite{campos2024conformal}. 2) \textit{Suggesting trustworthy solutions}. This involves informing humans about the model's internal workings, the reasoning, and confidence level behind its solutions \cite{boukhelifa2018evaluation, koyama2016selph, behrisch2014feedback}, through, for example, visual representations and verbal explanations \cite{heimerl2012visual, legg2013transformation, lai2019human}. 
    However, recent work suggests that LLM struggles to express genuine responses without deceit during the interaction \cite{huang2024concept}, violating the cooperative principle of sincerity. In a boarder context, without trustworthiness, LLMs risk undermining the foundation of long-term trust with humans \cite{sun2024trustllm}. This underscores the need for fostering trustworthy cooperation. 
\end{itemize}

Notably, recent advancements have explored enhancing model-assisted method with human-assisted method, leveraging human feedback to refine model performance \cite{lertvittayakumjorn2020find, ribeiro2022adaptive, tandon2022learning, li2023human, huang2024araida}. Some of them even empower the model to proactively seek human feedback \cite{huang2022inner, wang2023mint, cai2023human, mehta2023improving, zhang2024clamber}. This creates a dynamic loop where the human helps improve the model's decision-making process, potentially leading to better solutions. However, the ultimate decision-making authority remains with the human, placing this approach within the broader framework of model-assisted method.

\subsection{Triage-based Cooperation}
\label{tri}
\begin{center}
\begin{tcolorbox}[colback=gray!10,
                  colframe=black,
                  width=7.7cm,
                  arc=1mm, auto outer arc,
                  boxrule=0.5pt,
                  left=2pt, right=2pt, top=2pt, bottom=2pt
                 ]
\textbf{Triage-based Cooperation} refers to a cooperative process where tasks/data are strategically distributed between the human and the model based on their respective capabilities.
\end{tcolorbox}
\end{center}
\textbf{Overview}. Intuitively, humans excel at certain tasks while models demonstrate superiority in others, particularly repetitive and routine tasks \cite{fitts1951human}. This natural division of labor suggests an idea known as the "\textit{Humans Are Better At/Machines Are Better At}" (HABA-MABA) \cite{press1971toward, bradshaw2012human, dearden2000allocation}, which we term triage-based cooperation. 
Unlike the potential subordinate relationship observed in sequential cooperation, triage-based cooperation embraces an equal-partnership framework, taking responsibility for the tasks at hand and hence promoting a more balanced and cooperative dynamic.
To date, this approach has proven effective across diverse domains, including data annotation \cite{li2023coannotating, huang2024selective}, conversational information retrieval evaluation \cite{huang-etal-2023-reduce}, dialogue evaluation \cite{zhang2021human} and human-agent cooperation \cite{feng2024large}.

\begin{figure}[tb]
    \centering
    \setlength{\abovecaptionskip}{1pt}   
\setlength{\belowcaptionskip}{1pt}
    \includegraphics[width=0.38\textwidth]{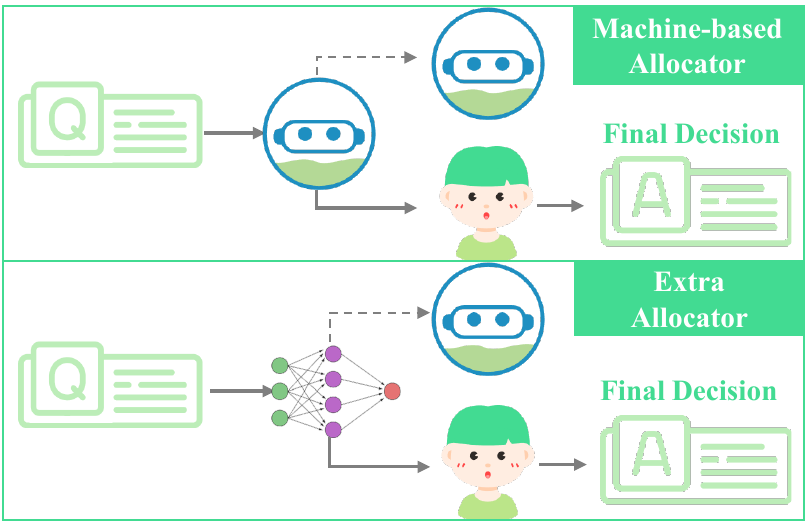}
    \caption{Triage-based cooperation allocates tasks based on capabilities of two parties, achieved by two types of allocators.
    It follows equal-partnership framework where cooperators are responsible for their own tasks.}
    \label{fig:tri}
    \vspace{-3mm}
\end{figure}

\textbf{Methods}. One of the challenges in triage-based cooperation is figuring out the best way to divide tasks. This requires accurately assessing the capabilities of both parties.
Additionally, modeling human capabilities adds another layer of difficulty, as it involves understanding the intricate interplay of various factors such as emotions \cite{vastfjall2016arithmetic} and self-control \cite{boureau2015deciding}. Therefore, current approaches primarily focus on evaluating the model's capabilities and seamlessly handing over tasks beyond its scope to the human \cite{raghu2019algorithmic, okati2021differentiable, madras2018predict}. This can be achieved through two primary allocators, illustrated in Figure \ref{fig:tri}. 
\begin{itemize}[leftmargin=*, itemsep=-4pt]
    \item \textbf{Model-based allocator}. It seamlessly integrates triage as an additional class within the model's decision-making process \cite{cortes2016learning, mozannar2020consistent}. For example, in a binary classification task, the model's output can include a third category representing \textit{triage-to-human}. This approach implicitly estimates the model's capabilities by leveraging a triage-aware cross-entropy loss \cite{thulasidasan2019combating}, making it well-suited for classification tasks. A possible way to extend this to generation tasks involves incorporating special tokens in the generated output to signal specific task information \cite{deng-etal-2022-pacific}, such as "\textit{[ToHuman] Why}" or "\textit{[ToModel] Response}". 
    
    \item \textbf{Extra allocator}. It acts as a filter, explicitly assessing the model's capabilities. It leverages metrics like prediction uncertainty \cite{ni2019calibration, li2023coannotating, ein-dor-etal-2020-active, xiao-etal-2023-freeal}, prediction unreliability \cite{jiang2018trust}, estimated error rate \cite{huang2024selective}, and data hardness \cite{de2020regression, wang2021classification, huang-etal-2023-reduce} to gauge the model's ability to handle a specific task. Implementation-wise, the extra allocator can be realized through automated evaluation metrics or any neural network. For example, \citet{huang2024selective} utilize a MLP to calculate the probability of model errors, while \citet{huang-etal-2023-reduce} employ ChatGPT to estimate data hardness. Usually, If the model's performance falls below a predefined threshold, the task is automatically handed over to a human. However, determining this threshold can be tricky. To this end, \citet{wilder2021learning} propose a dynamic threshold that automatically considers the uncertainty of both the model's predictions and the allocator's assessment. 
\end{itemize}

Notably, in this form of cooperation, cooperators adhere to the cooperative principle of Relation, providing task-relevant outputs within a strictly defined division of labor. The model is solely responsible for making predictions, and the human is expected to blindly accept those predictions for any samples the model chooses not to triage to a human. Crucially, there's no feedback loop in this cooperation: the model doesn't receive any human input or corrections during the process. This setup effectively creates a scenario where the two parties operate independently, each handling their assigned tasks without direct interaction.

\subsection{Joint Cooperation}
\label{joi}
\begin{center}
\begin{tcolorbox}[colback=gray!10,
                  colframe=black,
                  width=7.7cm,
                  arc=1mm, auto outer arc,
                  boxrule=0.5pt,
                  left=2pt, right=2pt, top=2pt, bottom=2pt
                 ]
\textbf{Joint Cooperation} refers to a cooperative process where both parties actively participate in the decision-making process, with the final outcome resulting from their collective decisions.
\end{tcolorbox}
\end{center}

\textbf{Overview}. Joint cooperation distinguishes itself from the sequential cooperation and triage-based cooperation by typically combining the outputs of both parties to yield a better one. While both joint operation and triage-based operation adhere to the equal-partnership framework, joint cooperation uniquely prioritizes a cooperative decision-making process where the human and the model work together on a single shared task. For better understanding, joint cooperation is founded on the recognition that humans and models excel in different ways and make distinct types of errors \cite{rosenfeld2018totally, geirhos2020beyond}. This diversity partially stems from their access to unique information, making joint cooperation that combines their perspectives particularly powerful for achieving more accurate and robust outcomes. 

\begin{figure}
    \centering
    \setlength{\abovecaptionskip}{1pt}   
\setlength{\belowcaptionskip}{1pt}
\includegraphics[width=0.49\textwidth]{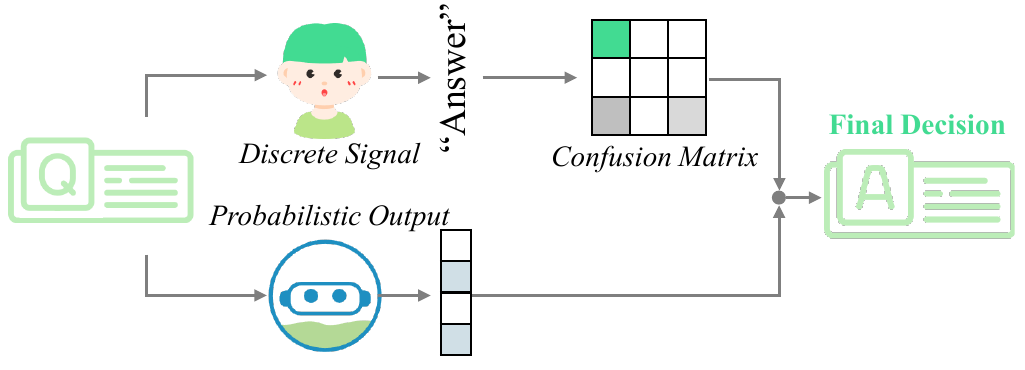}
    \caption{{\small Joint cooperation combines outputs of both parties. Both parties take responsibility for the final outputs.}}
    \label{fig:joi}
    \vspace{-5mm}
\end{figure}

\textbf{Methods}. Despite the potential benefits, combining the outputs of humans and models poses significant challenges. This is due to the inherent differences in their output formats: probabilistic model output and discrete human signals\footnote{Humans have high stochasticity in the expression of their uncertainty \cite{berkes2011spontaneous, orban2016neural}.}. To this end, current research, as illustrated in Figure \ref{fig:joi}, aims to merge the outputs of two parties in a probabilistic manner, which fits the discrete decision of the human into the confusion matrix \cite{xu1992methods, kuncheva2014combining, kerrigan2021combining}, which is used to estimate the human decision confidence. This approach can be implemented through supervised learning, utilizing pre-collected labeled datasets to estimate the confusion matrix, or through unsupervised learning, employing the EM algorithm to estimate the matrix without ground truth \cite{kerrigan2021combining}. Recent advancement on legal document matching task further enhances the EM estimation by employing cluster prototypes that record historical human decisions on the task \cite{huang2024comatching}.

Notably, while significant progress has been made, current methods are predominantly limited to classification tasks. This lack of applicability to generative tasks significantly restricts the practical use of joint cooperation in more scenarios. Furthermore, while enabling effective communication between the human and the model is paramount for unlocking the full potential of joint cooperation, current efforts primarily focus on how to combine prediction results, neglecting the critical need for bi-directional communication. 
This oversight presents a crucial area for future research.

\section{Open Challenges and Discussions}
We close our survey by discussing some trends and challenges. We categorize these into the following two primary segments.

\subsection{Technical Considerations}
\label{build}
\textbf{Which cooperation form is better}? While numerous methods for human-model cooperation have been proposed, tailored for different applications, a standardized benchmark is lacking, hindering our ability to objectively compare their effectiveness. This lack of benchmark makes it difficult to answer the crucial question: Which cooperation form is preferred to achieve optimal task performance while minimizing costs? For example, triage-based cooperation emphasizes independent work by the human and model, limiting information exchange and potentially hindering overall task completion quality. While joint and sequential cooperation require greater human intervention than triage-based cooperation, this can lead to higher human interaction costs, and the overall benefits of these labor costs on promoting task performance remain unclear.
To answer these questions, a comprehensive empirical benchmark is urgently needed. This may be achieved by, for example, conducting experiments on diverse NLP tasks using various user simulators \cite{wang-etal-2023-rethinking-evaluation, kocmi2023large, tang2023not, huang-etal-2023-reduce} to achieve a comprehensive assessment of the strengths, weaknesses, and potential risks associated with different forms of cooperation. 

\textbf{Human Uncertainty Estimation}. Uncertainty often plays a significant role in carrying out tasks in human-model cooperation. For example, accurate uncertainty estimation could improve task allocation in triage-based cooperation. However, human decision-making often lacks explicit uncertainty measurement \cite{yang2022optimal, kendall2017uncertainties, cha2021human, oh2020crowd}, suffering from \textit{epistemic uncertainty}\footnote{It pertains to uncertainty stemming from a lack of knowledge or information \cite{bland2012different}, i.e., "\textit{we are uncertain because we lack understanding}."}. Technically, human uncertainty estimation is a significant challenge. Existing approaches, such as ensemble learning \cite{raghu2019direct} or requiring the human to elicite uncertainty intervals \cite{zhang2021understanding, maadi2021collaborative}, are often unreliable due to the stochastic nature of human brain in expressing uncertainty \cite{orban2016neural, berkes2011spontaneous}. Moreover, simulating human decision-making, while promising, requires extensive human decision data \cite{ma2023should, bourgin2019cognitive}, making it less practical. Therefore, a greater focus on human uncertainty estimation is critical.

\textbf{Model Coordination}. Recent studies have found that LLMs often exhibit a one-size-fits-all strategy, struggling to adapt their behavior to diverse users without prior coordination \cite{chen2024style, huang2024concept}. This highlights the need for building a model with post-hoc coordination. 
However, retraining the model for each user is impractical. The challenge lies in building models that can seamlessly integrate into the existing workflows of any user, even those previously unseen by the model. However, training models to interact with humans effectively is inherently difficult, as it's often impossible to jointly train humans and models to coordinate their actions. Promising techniques may be the population-based training, which leverages diverse populations to improve the generalization ability of cooperative agents \cite{chen2024style, zhang2024strength, charakorn2020investigating}. An alternative approach involves developing self-evolving models \cite{tao2024surveyselfevolutionlargelanguage}, which hold the potential to autonomously acquire, refine, and learn from experiences gained through interactions with both humans and the surrounding environment.

\textbf{Cooperative Principles \& Behaviors}. While existing research often focuses on the rational behavior of both humans and models in human-model collaboration (i.e., both parties strive for optimal outcomes based on their objectives, rather than engaging in arbitrary or harmful actions), it lacks a clear articulation of the cooperative principles guiding the design of these interactions. Nevertheless, we've observed literature suggesting that LLMs may exhibit behaviors during the cooperation process that deviate from cooperation principles. This includes instances of deception \cite{huang2024concept}, task-irrelevant responses \cite{10.5555/3666122.3668386}, and over-loaded responses \cite{wang2023large}. Based on these findings, it is crucial to develop more fine-grained cooperation forms and behaviors grounded in these principles.

\subsection{Social Impact}
\label{use}
We underscore the practical value of the cooperation, extending beyond mere technical aspects.

\textbf{Trust Issue}. A common problem is trust calibration \cite{punzitowards}, where humans either under-rely or over-rely on the model's outputs. This can lead to misinterpretations, flawed decisions, and even amplified biases in fairness-related tasks. Even humans can fall victim to these issues, misjudging or misinterpreting model results due to misunderstandings or inappropriate reliance on the model's suggestions \cite{lai2019human, englich2006playing}. To foster trust and improve cooperation, explainable models are crucial, especially in tasks requiring ethical and unbiased outcomes. However, the challenges extend beyond just understanding the model. The issue of \textit{irony of automation} \cite{bainbridge1983ironies} also matters. Humans may misuse, disuse, or even abuse automation due to a lack of experience or understanding of its limitations. This is particularly true for non-professionals, who may have unrealistic expectations of the system's capabilities.

\textbf{Regulation \& Accountability}.
We emphasize the need for careful consideration of ethical, regulatory, and risk management aspects as human-model interaction becomes increasingly commonplace. 
The increasing accessibility of these models raises crucial ethical concerns, particularly regarding public acceptance of autonomous agents \cite{zlotowski2017can} and the need to address security vulnerabilities like hacking \cite{ferreira2019persuasion, chen2018description, zhang2024llms}. However, despite exploring various formalizations of human-model cooperation, the question of accountability in practical applications remains complex. Notably, the line of responsibility is blurred, making it difficult to determine who is ultimately responsible for the cooperation system's actions, especially when dealing with the equal-partnership framework where the model contributes equally as the human. This ambiguity highlights the need for clearer regulatory frameworks and ethical guidelines to ensure the responsible and accountable use of human-model cooperation systems.

\section{Conclusions}
\vspace{-1mm}
Intelligent models are expected to cooperate effectively within society for maximum productivity. In the era of LLMs, the moment has arrived to emphasize the advancement of human-model cooperation.
While numerous methods for human-model cooperation have emerged, information on how to formalize a human-model team is rather under-specified and scattered. To this end, this survey takes a crucial first step towards understanding human-model cooperation by offering a comprehensive overview of its definition, principles, and formalizations. We also introduce a novel taxonomy to categorize existing research, identifying key research frontiers and their associated challenges. With our survey, we provide a foundation for future exploration and pioneering advancements.


\section*{Limitations}
\paragraph{Multi-party Human-Model Cooperation} Human-model cooperation holds immense potential, but its complexity cannot be underestimated. While this survey focuses on the cooperation between a single human and a single model, real-world scenarios often involve multiple cooperators. These multi-party cooperation can involve a mix of different cooperation forms, e.g., triage-based and sequential cooperation, leading to intricate dynamics. Furthermore, human-human and model-model cooperation may also emerge within these teams, creating further layers of complexity. Instead, we chose to begin with a more simplified scenario (i.e., a single human and a single model), with the aim of bringing together the under-specified and scattered information about how to formalize an effective human-model cooperation to achieve collective outputs. Given the scope of our work, we will leave the exploration of multi-party human-model cooperation for future research.

\paragraph{Human-Model Collaboration} Our survey solely focuses on human-model cooperation, excluding human-model collaboration. Human-model collaboration goes beyond the cooperation; it calls for the introduction of bidirectional communication and co-decision making that harnesses the potential of both human and model capabilities \citep{punzitowards}. However, current methods on sequential cooperation, while effectively leveraging the language capabilities of LLMs for communication, such as in conversational information seeking \cite{zamani2022conversational}, fall short in facilitating the collaborative decision-making essential for true human-model collaboration. Additionally, methods on triage-based and joint cooperation often neglect the crucial aspect of communication between the two parties. Notably, establishing professional communication is paramount, as it allows both parties to validate their rationality, recognize each other's limitations, and engage in a reciprocal learning process \cite{rabinowitz2018machine}. However, while significant progress has been made in exploring the cooperation, a substantial gap persists between these concepts and the practical implementation of human-model collaboration systems. 

\paragraph{Human-Model Non-cooperation} 
Beyond the cooperation, models can also engage in non-cooperative interactions with the human \cite{zhang2024strength, deng2023prompting, deng2023plug}. They can negotiate prices with users, employing strategic tactics to reach a favorable outcome \cite{he2018decoupling}. They can also attempt to persuade users to donate to charitable causes, leveraging their linguistic prowess to sway opinions and evoke generosity \cite{wang-etal-2019-persuasion}. Considering the large scope of human-model non-cooperation, we will dedicate future research to exploring this area in greater depth.

\section*{Acknowledgments}
This work was supported in part by the National Natural Science Foundation of China (No. 62272330 and No.U24A20328); in part by the Fundamental Research Funds for the Central Universities (No. YJ202219); in part by the Science Fund for Creative Research Groups of Sichuan Province Natural Science Foundation (No. 2024NSFTD0035); in part by the National Major Scientific Instruments and Equipments Development Project of Natural Science Foundation of China under Grant (No. 62427820); in part by the Natural Science Foundation of Sichuan (No. 2024YFHZ0233); in part by the Singapore Ministry of Education (MOE) Academic Research Fund (AcRF) Tier 1 grant (No. MSS24C012).

\bibliography{custom}

\appendix

\section{Systematic Literature Review Process}
\label{liter}
This systematic review examines a broad range of research in Natural Language Processing (NLP), Human-Computer Interaction, and Machine Learning, with a particular interest on the groundbreaking advancements in Large Language Models (LLMs). We introduce the details below.
\begin{itemize}[leftmargin=*]
    \item \textbf{Identification with Keywords}. Our research began with a collection of papers published in the fields of Natural Language Processing and Machine Learning. To achieve this, we applied a rigorous keyword-based screening process using a carefully curated list of keywords and their combinations and variations. To expand our understanding of this literature, we then gathered the papers that cited these selected papers, as well as the papers they cited themselves. This resulted in a final set of 1007 papers. Our keywords include \textit{human-model, human-AI, human-model, human-agent, cooperation, collaboration, interaction/interactive, teaming, co-decision making, human-in-the-loop, machine-in-the-loop, algorithm-in-the-loop, active learning, conversational information seeking, conversational search, conversational recommender system, interactive machine learning, interactive NLP, dialogue system}, and \textit{human feedback}.
    \item \textbf{Inclusion and Exclusion Criteria}. We narrowed down our initial set of 1007 papers based on carefully defined inclusion and exclusion criteria. These criteria were directly aligned with our core research question: \textit{how to formalize an effective human-model cooperation to achieve collective outputs}. We selected papers that offered potential answers to this question, while excluding those that did not meet our criteria. This process resulted in a corpus of 147 papers for further analysis.
\end{itemize}

\section{Related Survey}
\label{related}
The pursuit of AI systems capable of interacting and cooperating effectively with humans has long been a central focus of research. While numerous surveys have explored this topic from various perspectives, such as Machine Learning and Human-Computer Interaction \cite{fails2003interactive, bland2012different, zhang2021ideal, dellermann201sd9hybrid, donahue2022human, wondimu2022interactive, punzitowards}, recent advancements in deep learning and large language models have spurred significant interest in human-model/human-model cooperation within NLP community \cite{wang2023interactive}. This leads to a growing body of dedicated NLP surveys \cite{wang2021putting, wan2022user, wang2023interactive, wu2023designing, yang2024human, xi2023rise, gao2024taxonomy}. However, those surveys focus more on introducing key elements of cooperation, such as cooperation interfaces, message understanding and utilization, and applications to NLP tasks. In particular, \citet{gao2024taxonomy} propose a comprehensive framework encompassing four key phases – planning, facilitating, iterating, and testing. and develop a taxonomy of four primary interaction modes: Standard Prompting, User Interface, Context-based, and Agent Facilitator. Additionally, \citet{wang2021putting} summarize the NLP tasks, goals, types of human feedback and user interfaces of exiting human-model cooperation methods, while \citet{wang2023interactive} delve deeper into the intricacies of cooperation systems, focusing on the characteristics of interactive objects, interaction interfaces, and message fusion strategies. Furthermore, \citet{wu2023designing, yang2024human} provide practical insights through four case studies, demonstrating best practices for evaluating human-model cooperation systems, designing user-friendly interfaces, and leveraging human interaction to enhance the performance of NLP models. Finally, \citet{xi2023rise} explore the diverse landscape of human-Agent collaboration, drawing upon existing research to re-examine the distinct roles assumed by humans and agents during cooperative endeavors. This exploration builds upon the framework proposed by \cite{zhang2021ideal}, which distinguishes between two primary models: the assistor-executor and the equal-partnership frameworks. 
While significant strides have been made in understanding human-model interaction, current research largely focuses on sequential cooperation and neglects the exploration of fundamental principles and formalizations for achieving effective human-model collaboration. Existing work, such as that by \citet{wan2022user}, try to formalize different types of human-model interactions but remain limited in scope, focusing on only sequential cooperation setting. This gap highlights the need for a more comprehensive and systematic analysis of the underlying principles and formalizations governing human-model cooperation.
In this paper, we take the first step to present a thorough review of human-model cooperation, exploring its principles, formalizations, and open challenges. Further, we introduce a new taxonomy that provides a unified perspective to summarize existing approaches on enabling effective human-model cooperation in decision-making.

\section{Formal Definition of Human-Model Cooperation}
This section aims to enables a formal perspective towards the human-model cooperation. 

\textbf{Notations}. Considering a task $T$ and a 2-participant interaction, we define the participants set as $N=\{H, M\}$, where $H$ denotes the human participant and $M$ denotes the model participant. Additionally, 
the decision or action variable of participant $i \in N$ is represented as $a_i\in A_i$, where $A_i$ is the action set\footnote{This set could be finite, meaning that the human or the model has only a finite number of possible actions, or it could be infinite but finite-dimensional (e.g., the unit interval, $[0, 1]$), or even infinite-dimensional (e.g., the space of all continuous values on the interval $[0, 1]$).} of participant $i$. 
We can express the 2-tuple of action variables of all players as $a=(a_H, a_M)$. 
Introducing possibly coupled constraints, let $\Omega = \{\Omega_H, \Omega_M\} \in A$ be the constraint set that records the interaction-oriented or task-oriented constraints for human-model interaction, where $A=A_H \times A_M$. For example, a constraint could demand actions that result in high time efficiency or reduced cognitive workload \cite{zhang2021human}. Therefore, for a 2-tuple of action variables $a$ to be feasible, it is necessary for $a \in \Omega$. 
Additionally, each participant $i$ have a policy $\pi_i$ that returns the probability distribution over the set of possible actions $A_i$. 

\textbf{Formal Definition}. Human-model cooperation involves the human and the model working together as a unified team, engaging in the decision-making process of shared tasks to achieve a \textit{shared goal}. This shared goal could be measured by a shared utility function $U$ (e.g., the task success rate, or an individual’s subjective evaluation of the desirability or satisfaction) when both the human and the model act according to their respective policies. In this case, the primary objective of human-model cooperation could be formalized as the following optimization problem:
\begin{center}
\begin{tcolorbox}[colback=gray!10,
                  colframe=black,
                  width=7.6cm,
                  arc=1mm, auto outer arc,
                  boxrule=0.5pt
                 ]
\begin{equation}\nonumber
\begin{split}
    \max_{a_i} \max_{a_{-i}} \quad &U(a_i, a_{-i})\\
    s.t. \quad &a_i \in \Omega_i, i \in N\\
    &a_i \sim \pi_i(a|a_{-i}, T, \Omega, G_i),
\end{split}
\end{equation}
where $a_{-i}$ stands for the action variable of other participant except $i$.
When the both parties work independently, we have $a_i \sim \pi_i(a|T, \Omega, G_i)$. 
\end{tcolorbox}
\end{center}

\section{Overview of the Survey}\label{app:overview}
\begin{figure*}
    \centering
    \includegraphics[width=0.99\textwidth]{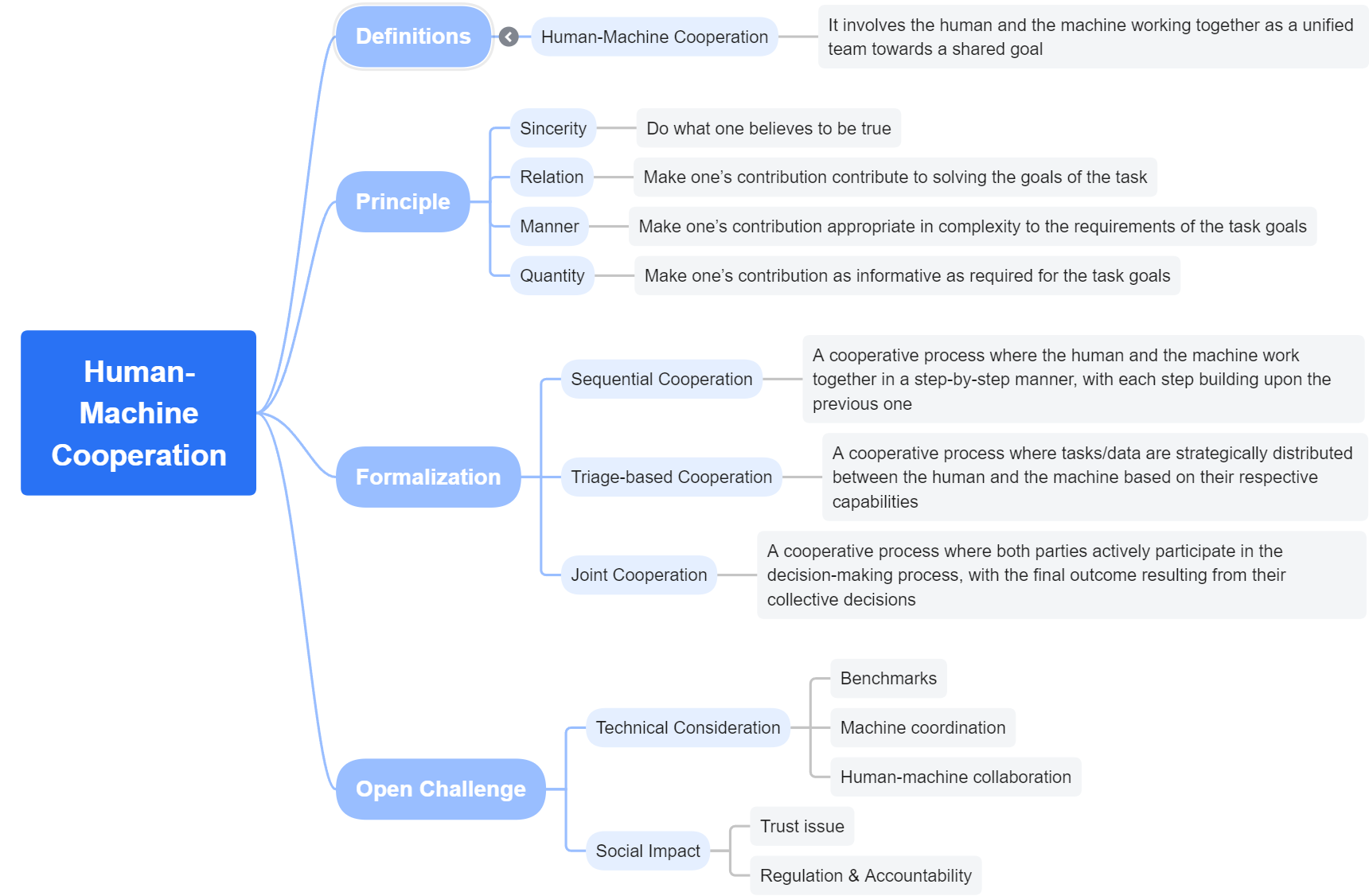}
    \caption{Overview of the survey.}
    \label{fig:overvi}
\end{figure*}

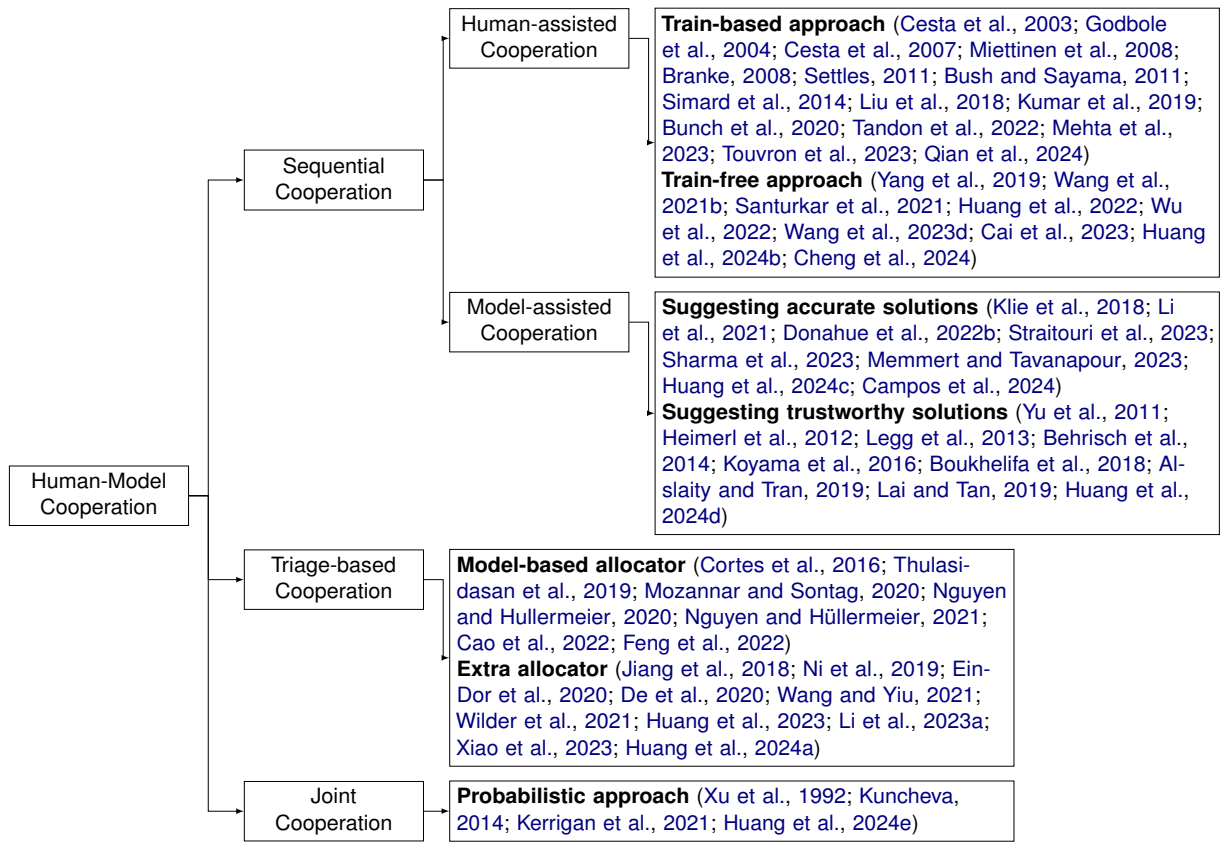
\begin{figure*}
    \centering
    
\tikzset{
    basic/.style  = {draw, text width=3cm, align=center, font=\sffamily, rectangle},
    basic1/.style  = {draw, text width=10cm, align=left, font=\sffamily, rectangle},
    root/.style   = {basic, rounded corners=2pt, thin, align=center, fill=green!30},
    onode/.style = {basic, thin, rounded corners=2pt, align=center, fill=green!60,text width=3cm,},
    tnode/.style = {basic, thin, align=left, fill=pink!60, text width=15em, align=center},
    xnode/.style = {basic, thin, rounded corners=2pt, align=center, fill=blue!20,text width=5cm,},
    wnode/.style = {basic, thin, align=left, fill=pink!10!blue!80!red!10, text width=6.5em},
    edge from parent/.style={draw=black, edge from parent fork right}

}
\resizebox{\textwidth}{!}{%
\begin{forest} 
for tree={
    grow=east,
    growth parent anchor=west,
    parent anchor=east,
    child anchor=west,
    edge path={\noexpand\path[\forestoption{edge},->, >={latex}] 
         (!u.parent anchor) -- +(10pt,0pt) |-  (.child anchor) 
         \forestoption{edge label};}
}
[Human-Model Cooperation, basic, l sep=10mm,
        [Joint\\ Cooperation, basic, 
          [\textbf{Probabilistic approach} \cite{xu1992methods, kuncheva2014combining, kerrigan2021combining, huang2024comatching}, basic1]
        ]
        [Triage-based\\ Cooperation, basic,
          [\textbf{Model-based allocator} \cite{cortes2016learning, thulasidasan2019combating, mozannar2020consistent, nguyen2020reliable, nguyen2021multilabel, cao2022generalizing, feng2022towards} \\
          \textbf{Extra allocator} \cite{jiang2018trust, ni2019calibration, ein-dor-etal-2020-active, de2020regression, wang2021classification, wilder2021learning, huang-etal-2023-reduce, li2023coannotating, xiao-etal-2023-freeal, huang2024selective}
          , basic1]
        ]
        [Sequential\\ Cooperation, basic
          [Model-assisted\\ Cooperation, basic, 
              [\textbf{Suggesting accurate solutions} \cite{klie-etal-2018-inception, li2021fitannotator, donahue2022pick, straitouri2023improving, sharma2023human, memmert2023towards, huang2024araida, campos2024conformal} \\
              \textbf{Suggesting trustworthy solutions} \cite{yu2011toward, heimerl2012visual, legg2013transformation, behrisch2014feedback, koyama2016selph, boukhelifa2018evaluation, alslaity2019towards, lai2019human, huang2024concept}, basic1]
          ]
          [Human-assisted\\ Cooperation, basic, 
            [\textbf{Train-based approach} \cite{cesta2003csp, godbole2004document, cesta2007mexar2, miettinen2008introduction, branke2008consideration, settles2011closing, bush2011hyperinteractive, simard2014ice, liu2018dialogue, kumar2019didn, bunch2020human, tandon2022learning, mehta2023improving, touvron2023llama, qian2024tell}\\
            \textbf{Train-free approach} \cite{yang2019study, wang2021gam, santurkar2021editing, huang2022inner, wu2022ai, wang2023mint, cai2023human, huang2024dreditor, cheng2024editing}, basic1]
          ]
        ]  
]
\end{forest}
}
    \caption{Literature survey tree. We list more representative methods for different categories of human-model cooperation methods.}
    \label{fig:lit_surv}
\end{figure*}

We present the overview of this survey in Figure \ref{fig:overvi} and the literature survey tree in Figure \ref{fig:lit_surv}. This visually illustrates the scope and structure of our research, offering a clear understanding of the key areas covered in our survey.

\section{History of Cooperation Forms}
This section aims to offer a brief overview of the history and evolutionary of different cooperation forms. In particular, in terms of cooperative complexity, sequential and triage-based cooperation are generally considered simpler than joint cooperation. This is reflected in their historical development, with the former two emerging as early as the 1970s \cite{press1971toward, wallenius1975comparative} and continuously becoming two mainstream forms, while joint cooperation has only gained prominence in recent years \cite{kerrigan2021combining, huang2024comatching}. This shift is attributed to significant technological advancements that have enabled the seamless integration of human and model decision-making algorithms, paving the way for more sophisticated joint cooperative models. As for the sequential cooperation, it has consistently been a focal point of research in human-model cooperation, whose representatives include human-in-the-loop \cite{wang2021putting} and machine-in-the-loop \cite{green2020algorithm} approaches. As for the triage-based cooperation, research community start this research topic since 1970s \cite{chow1970optimum, hellman1970nearest}, where the model has the option to abstain from making a prediction when they are likely to make a mistake. This approach gained further traction in the 2000s \cite{hendrickx2024machine}, highlighting the importance of allowing models to acknowledge their limitations. More recently, the NLP community, starting around the 2020s \cite{zhang2021human, feng2024large}, has embraced this form of cooperation, recognizing that LLMs and LMs are not omnipotent and that humans and models possess complementary strengths.

\section{Why Human-Model Cooperation Matters}
\textbf{Overall}. Both humans and models possess unique strengths and weaknesses. While models excel at processing vast amounts of explicit knowledge (i.e., formalized and codified knowledge like documents), humans often possess crucial insights that are difficult to codify for models \cite{dellermann201sd9hybrid}. For example, a judge gains valuable knowledge about a defendant through interaction, a skill that current models struggle to replicate. This inherent complementarity between human and model intelligence motivates the development of collaborative systems, where each party contributes their unique strengths to enhance task performance.

\textbf{Humans possess human intelligence}. Human intelligence is a multifaceted and dynamic aspect of human nature, encompassing a broad spectrum of mental abilities and skills. It involves the capacity to learn, understand complex concepts, and apply logic and reasoning to solve problems. Furthermore, human intelligence is intrinsically linked to motivation and self-awareness, as individuals with high intelligence often demonstrate a strong drive to achieve their goals and possess a deep understanding of their own thoughts, emotions, and behaviors. On the contrary, models lack essential human capabilities required to attain human-level general intelligence \cite{ma2023brain}, such as logic reasoning \cite{xu2023large} and causal reasoning \cite{jin2023can}. While GPT-4 and o1 represent significant advancements in model capabilities, it remains uncertain whether they will ever fully eclipse human intelligence.

\textbf{Humans are rich in tacit knowledge}, encompassing practical expertise that enables them to perform specific actions \cite{ryle1945knowing}. In contrast to formalized, codified, or explicit knowledge, tacit knowledge (or implicit knowledge) is information that is challenging to articulate and is embedded in individual experiences in forms such as motor skills, personal wisdom, intuitions and hunches. Consequently, it is difficult to convey to others through writing or verbalization, and is only evident in human behaviors \cite{dellermann201sd9hybrid}. While tacit knowledge has been examined in various fields \cite{anshari2023optimisation, muller2019data}, integrating it into AI models has always posed a challenge. In this context, involving human experts directly in the model decision-making process through interactions offers a promising approach to incorporate human tacit knowledge, ultimately enhancing the effectiveness of tasks performed by the human-model team.

\textbf{Social Impact}. Ultimately, NLP models are designed to serve human needs. However, the lack of human involvement in model predictions can lead to a lack of trust in its outcomes \cite{punzitowards}, as previously observed. This challenge is particularly acute in high-stakes domains, where trust in AI is paramount. Consider, for example, a doctor's reliance on an LLM's diagnosis. The potential for skepticism is amplified by the fact that LLMs often lack clear explainability and reliable uncertainty quantification \cite{gawlikowski2023survey}, raising concerns about their reliability and transparency. This fundamental issue presents a significant barrier to wider adoption of LLMs in critical applications. In this case, it's essential to involve humans in the decision-making process through effective human-model cooperation. This can be achieved by presenting humans with multiple options for review or by assigning models to simpler tasks (like diagnosing colds) while humans handle more complex ones (like diagnosing depression).

\section{Typical Applications of Human-Model Cooperation}
\label{applica}
This section explores the practical applications of human-model cooperation, including data annotation and information seeking. Each of these areas exemplifies how humans and models can work together effectively, leveraging their unique strengths to achieve common goals. By examining these examples, we can gain a better understanding of the diverse ways in which humans and models can complement each other, leading to innovative and impactful outcomes.

\subsection{Human-Model Cooperation for Data Annotation}
Data annotation is a challenging (and somehow trivial) task in managing the trade-off between data quality and human resource investment. To touch upon this challenge, existing studies resort to human-model cooperation as follows:

\begin{figure}[!th]
    \centering
    \includegraphics[width=0.49\textwidth]{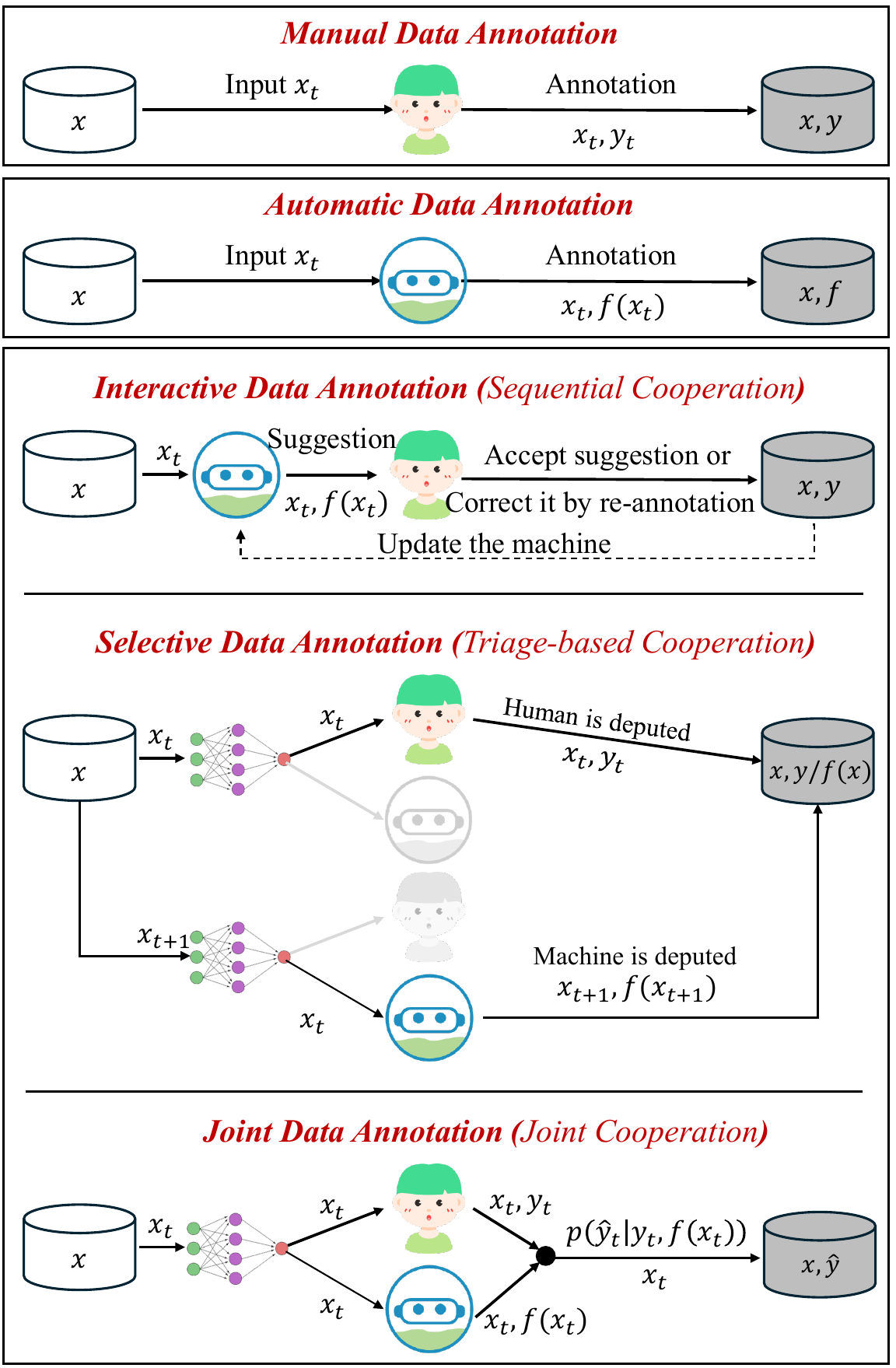}
    \caption{Typical applications of human-model cooperation in data annotation, including manual data annotation, automatic data annotation, and human-model data annotation.}
    \label{fig:enter-labeldata}
\end{figure}

\textbf{\textit{Interactive Data Annotation} --Sequential Cooperation}. Interactive data annotation, also known as semi-automatic annotation, streamlines the labeling process by introducing an annotation model that proposes suggestions (model annotations, denoted as $f(x)$) to a human annotator. As illustrated in Figure~\ref{fig:enter-labeldata}, the human annotator reviews each suggested label and either accepts it if correct or manually corrects it. Compared to traditional manual annotation, this interactive approach significantly reduces human effort by eliminating the need to generate labels from scratch. Research by \cite{klie-etal-2018-inception,klie-etal-2020-zero,le2021interactive, huang2024araida} has shown the potential for interactive annotation to accelerate data labeling. The annotation process concludes when all available unlabeled data has been reviewed and annotated.

\textbf{\textit{Selective Data Annotation} -- Triage-based Cooperation}. In this scenario, the data annotation process can be conceptualized as a data triage problem. A specialized  data triage module dynamically assigns data points to either a human annotator or an annotation model on the fly. This module generates a binary triage signal (0 or 1) to determine the annotation responsibility. The annotation process concludes upon the exhaustion of available resources or the successful annotation of all data. This dynamic approach, as explored by \citet{li2023coannotating, ein-dor-etal-2020-active, huang2024selective}, optimizes the collaboration between human expertise and model learning capabilities in data annotation.

\textbf{\textit{Joint Data Annotation} -- Joint Cooperation}. There are mainly two reasons for limited work on data annotation in this setting.  Firstly, integrating discrete human annotations with probabilistic model annotations is complex. Secondly, this form of collaboration necessitates continuous human involvement, making it expensive. Despite these obstacles, certain task-specified algorithms demonstrate the potential for combining human and model outputs, particularly for tasks where human cost is acceptable. Notably, \citet{huang2024comatching} focus on legal case matching, a domain where legal professionals prioritize accuracy and are willing to invest significant effort. Their approach involves both legal practitioners and models cooperatively annotating key sentences within lengthy legal documents. A probabilistic method then combines these annotations, enabling a joint human-model decision-making process that minimizes errors. 

\begin{figure}[!th]
    \centering
    \includegraphics[width=0.49\textwidth]{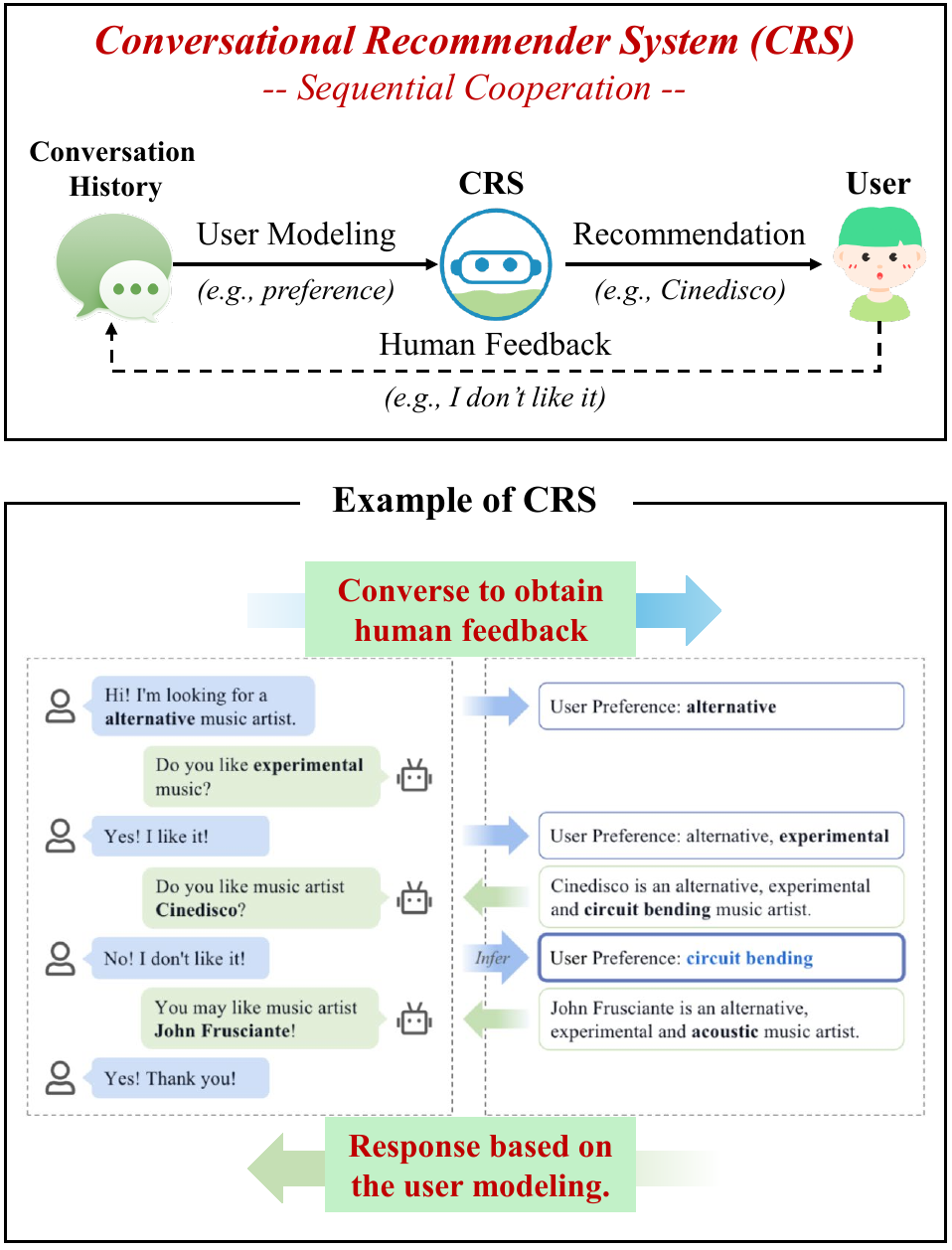}
    \caption{Typical applications of human-model cooperation in information seeking. One prominent example is the conversational recommender system (CRS), where the system and the user engage in sequential cooperation, effectively forming the assistant-executor framework. This dynamic interaction allows the system to assist the user in identifying preferred items by incorporating user feedback. Note that this figure draws inspiration from the work of \citet{lei2020interactive}.}
    \label{fig:enter-crs}
\end{figure}

\subsection{Human-Model Cooperation for Information Seeking}
\textbf{\textit{Conversational Information Seeking} -- Sequential Cooperation}. One prominent example is conversational information seeking (CIK), which involves an information retrieval system that utilizes a conversational interface to understand and adapt to users' dynamic preferences, providing real-time retrieval of information based on estimated user needs during (multi-turn) conversations \cite{zamani2022conversational, gao2021advances}. CIK allows models to gather relevant information about user behavior, intentions, and preferences through verbal communication. Specifically, CIK encompasses three primary research areas: conversational search \cite{zhang2018towards, rosset2020leading}, conversational recommender system \cite{lei2020interactive, huang2024concept}, and conversational question answering \cite{reddy2019coqa, peng2022godel}. 
Notably, information seeking inherently presents a clear division of labor between the human and the model, with a hierarchical structure where the model serves the human by retrieving desired information. This natural division of roles typically leads to sequential cooperation in conversational information seeking tasks, where the model act as an assistant. In this framework, the human guides the process by providing search queries or feedback, while the model leverages its computational power to efficiently locate relevant information. For better understanding, we illustrate the conversational recommender system (CRS) in Figure \ref{fig:enter-crs}. 



\end{document}